\def\year{2022}\relax
\documentclass[letterpaper]{article} 
\usepackage{aaai22}  
\usepackage{times}  
\usepackage{helvet}  
\usepackage{courier}  
\usepackage[hyphens]{url}  
\usepackage{graphicx} 
\usepackage{natbib}  
\usepackage{caption} 
\DeclareCaptionStyle{ruled}{labelfont=normalfont,labelsep=colon,strut=off} 
\usepackage{algorithm}
\usepackage{algorithmic}

\usepackage{newfloat}
\usepackage{listings}
\floatstyle{ruled}
\newfloat{listing}{tb}{lst}{}
\floatname{listing}{Listing}
\usepackage{amsmath}
\usepackage{booktabs}
\usepackage{multirow}
\usepackage{graphicx}
\DeclareMathOperator*{\argmax}{arg\,max}

\title{Overcome Anterograde Forgetting with Cycle Memory Networks}
\author{  Jian Peng \textsuperscript{\rm 1},
    Dingqi Ye \textsuperscript{\rm 1},
    Bo Tang\textsuperscript{\rm 2},
    Yinjie Lei \textsuperscript{\rm 3},
    Yu Liu \textsuperscript{\rm 4},
    Haifeng Li \textsuperscript{\rm 1,*}
}
\affiliations{

    \textsuperscript{\rm 1} School of Geosciences and Info-Physics, Central South University, Changsha 410083  \\
    \textsuperscript{\rm 2} Department of Electrical and Computer Engineering,
Mississippi State University, Starkville, MS 39762 USA  \\
    \textsuperscript{\rm 3} College of Electronics and Information Engineering,
Sichuan University, Chengdu 610017, China  \\
    \textsuperscript{\rm 4} School of Earth and Space Sciences, Institute of
Remote Sensing and Geographic Information System, Peking University,
Beijing 100871, China.  \\

}

\usepackage{bibentry}

\begin{document}

\maketitle

\begin{abstract}
Learning from a sequence of tasks for a lifetime is essential for an agent towards artificial general intelligence. This requires the agent to continuously learn and memorize new knowledge without interference. This paper first demonstrates a fundamental issue of lifelong learning using neural networks, named \textit{anterograde forgetting}, i.e., preserving and transferring memory may inhibit the learning of new knowledge. This is attributed to the fact that the learning capacity of a neural network will be reduced as it keeps memorizing historical knowledge, and the fact that the conceptual confusion may occur as it transfers irrelevant old knowledge to the current task. This work proposes a general framework named Cycled Memory Networks (CMN) to address the anterograde forgetting in neural networks for lifelong learning. The CMN consists of two individual memory networks to store short-term and long-term memories to avoid capacity shrinkage. A transfer cell is designed to connect these two memory networks, enabling knowledge transfer from the long-term memory network to the short-term memory network to mitigate the conceptual confusion, and a memory consolidation mechanism is developed to integrate short-term knowledge into the long-term memory network for knowledge accumulation. Experimental results demonstrate that the CMN can effectively address the anterograde forgetting on several task-related, task-conflict, class-incremental and cross-domain benchmarks.
\end{abstract}

\section{Introduction}
Humans can continuously acquire and transfer knowledge and skills over time. This ability, referred as lifelong learning, is also crucial to autonomous agents when they interact with dynamic real-world environments with non-stationary data and/or incremental tasks \cite{hadsell2020embracing,parisi2019continual}, yet given finite resources, e.g., computational power or storage capacity. To enable machines to possess such human-like lifelong learning capabilities, it requires agents to be able to learn and accumulate knowledge sequentially \cite{chen2018lifelong}. An ideal lifelong learning system is able to: \textit{(i)} continuously acquire new knowledge from new tasks \cite{lopez2017gradient}; \textit{(ii)} preserve previously learned knowledge in old tasks, i.e., overcoming catastrophic forgetting \cite{mccloskey1989catastrophic,Kirkpatrick2016Overcoming,pfulb2019comprehensive}; \textit{(iii)} transfer previous knowledge from the memory to facilitate the learning of new tasks \cite{swaroop2019improving,schwarz2018progress}; and \textit{(iv)} use fewer parameters and storage \cite{pfulb2019comprehensive,farquhar2018towards}.

\begin{figure}[h]
\centering
\includegraphics[width=\linewidth]{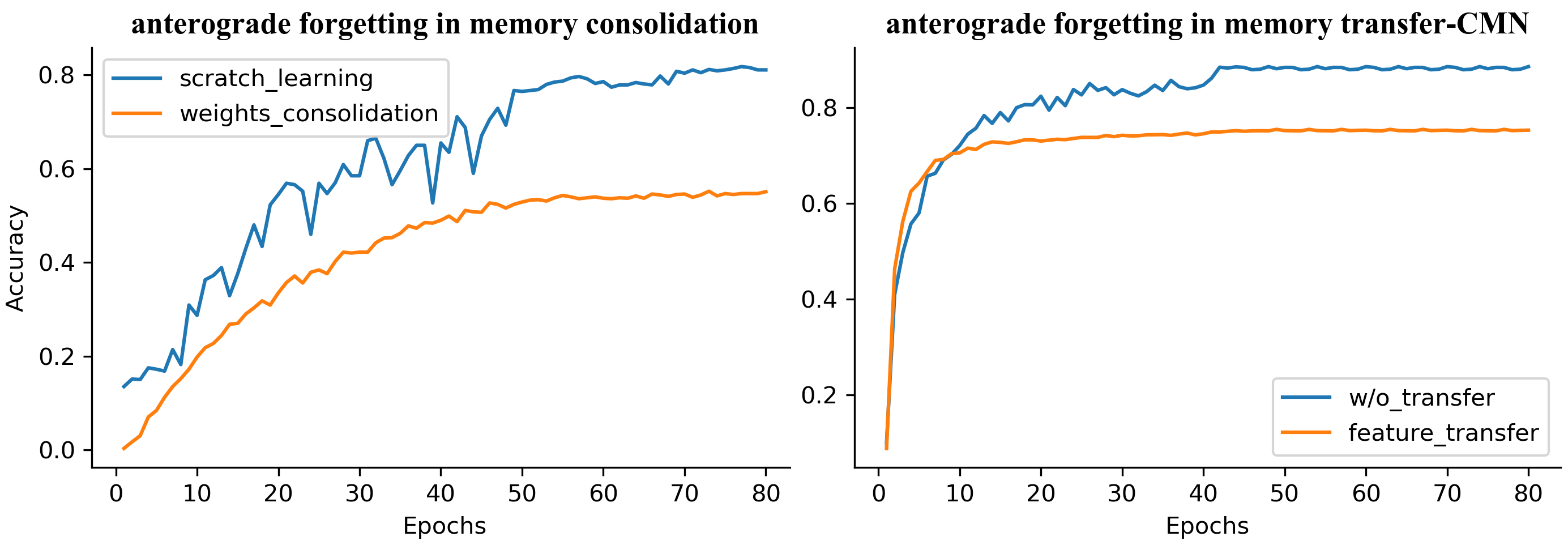}
\caption{Two scenarios of the anterograde forgetting issue in neural networks for lifelong learning. Preserving (\textit{left}) and transferring (\textit{right}) old knowledge could prevent the model from learning of new knowledge.
(\textit{{Left}}): The experiment was conducted on CIFAR-100 for the first two tasks. Models were trained on the first task and tested on the second task.  As a control, $scratch\_learning$ trained models from scratch without the influences of anterograde forgetting. The $weights\_consolidation$, i.e., EWC \cite{Kirkpatrick2016Overcoming}, is one weight consolidation method that adds a regularization item to remember the previous task's knowledge.
(\textit{Right}): Similarly, the experiment was performed on CIFAR-10 for the first two tasks. The $feature\_transfer$, i.e., Progressive Neural Networks \cite{rusu2016progressive}, is a feature transfer-based method. Compared with $w/o\_transfer$, it transfers features from the model of previous tasks to the current task's learning.}
\label{fig1}
\end{figure}

A number of lifelong learning research have been conducted over the past decades \cite{parisi2019continual,chen2018lifelong,mccloskey1989catastrophic,aljundi2018memory, Goodfellow2013An,robins1993catastrophic}. However, few studies can achieve all above capabilities of an ideal lifelong learning system. In particular, memory transfer has been long neglected. The association between memory and learning remains difficult for current lifelong learning models, leading to the \textit{anterograde forgetting}. In contrast to the catastrophic forgetting in which learning of new knowledge could eliminate old knowledge quickly, the anterograde forgetting describes the phenomenon that the old knowledge in the memory could prevent the learning of new knowledge. This issue could occur in both memory retention and memory transfer. On the one hand, retaining old knowledge continuously occupies model capacity left for learning new knowledge. Such capacity shrinkage hinders models from learning new knowledge adequately. On the other hand, transferring irrelevant old knowledge introduces noise, which leads to conceptual confusions while learning new knowledge. Our empirical study demonstrates these two scenarios of the anterograde forgetting, as shown in Fig. \ref{fig1}, where performance degradation of a lifelong learning model can be observed when it learns a new task with old knowledge: \textit{(i)} the model equipped with a memory consolidation algorithm converges slowly and performs poorly compared to a model trained without memory retention; and \textit{(ii)} the model with a memory transfer algorithm has a lower accuracy than that of the model without memory transfer. Note that two pairs of models that are compared have the same experimental setting, i.e., dataset, hyperparameters, and initial parameters, except for the difference of models with and without memory consolidation and memory transfer.

Why are humans capable of learning from a sequence of tasks without catastrophic and anterograde forgetting?
Neuroscience studies suggest that a short-term memory is first formed when learning a task and is subsequently transformed into a long-term memory \cite{trannoy2011parallel,dudai2015consolidation,izquierdo1999separate}. Relatively, long-term memory retrieves and reactivates similar memories and facilitates short-term memory formation \cite{hulme1991memory}.
Inspired by the above research, we introduce the following principles for the development of effective lifelong learning systems: \textit{(i)} two individual networks for encoding new and old knowledge separately to avoid conflicts, \textit{(ii)} an explicit mechanism to integrate new knowledge into old knowledge, and \textit{(iii)} specialized modules to transfer old knowledge to facilitate the learning of new knowledge.

\begin{figure*}[t]
\centering
\includegraphics[width=0.9\linewidth]{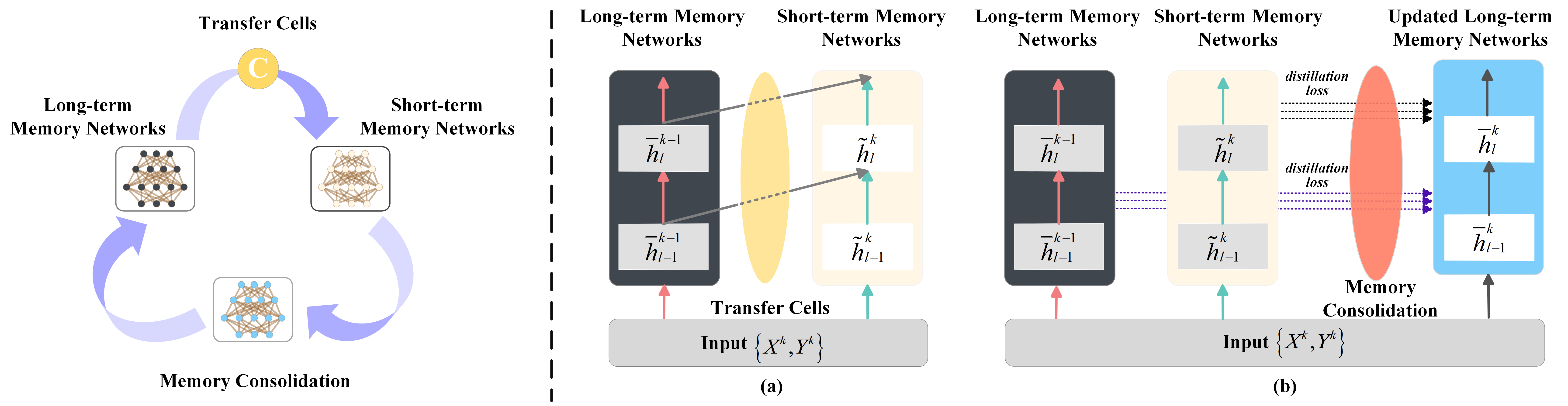}
\caption{An overview of the proposed CMN. (\textit{Left}): A memory-learning spiral loop to enable knowledge transformation and transfer between long- and short-term memory. (\textit{Right}): the architecture of CMN, (a) L-S transfer half-cycle, the process of long-term memory transferring to short-term memory via the transfer cells; (b) S-L transform half-cycle, the process of short-term memory transforming to the long-term memory via knowledge distillation.}
\label{fig2}
\end{figure*}

This work proposes a general lifelong learning framework based on the above principles, called \textit{Cycled Memory Networks (CMN)}, as shown in Fig. \ref{fig2}. It has a \textit{memory-learning spiral loop}: when the CMN learns a task, new knowledge will be extracted and stored in a short-term memory network, followed by a memory consolidation process which
integrates the short-term memory into a long-term memory network; when it continuously learns a new task in the future, a transfer cell is developed to effectively transfer the knowledge in the long-term memory network to assist learning new knowledge. Extensive experiments are conducted to evaluate the proposed framework on several benchmarks with known and unknown task relations. Experimental results show that the proposed CMN can overcome both anterograde and catastrophic forgetting issues, and achieve precise knowledge migration between long- and short-term memory networks. Specifically, the contribution of this paper can be summarized as follows,
\begin{itemize}
    \item  The anterograde forgetting issue is first demonstrated in lifelong learning.
    \item  A biologically inspired lifelong learning framework is developed through an association of memory and learning.
    \item  Two individual memory networks: short-term and long-term memory networks are introduced to separately store new and old knowledge, and a transfer cell is developed to learn the knowledge transferability from the long-term memory network to the short-term memory network.
    \item A memory consolidation mechanism is introduced to integrate the new knowledge learned in the short-term memory network into the long-term memory network to achieve an effective knowledge accumulation.
\end{itemize}

\section{Related Works}
\textbf{Memory consolidation for lifelong learning.} Lifelong learning, also known as continual learning \cite{ring1994continual} or sequential learning \cite{aljundi2018selfless}, has been recently receiving increasing attention in machine learning and artificial intelligence \cite{chen2018lifelong}. The majority of previous works focused on addressing the well-known catastrophic forgetting issue \cite{mccloskey1989catastrophic}. According to the mechanism of memory consolidation, current approaches are categorized into three types: \textit{(i)} Experiential rehearsal-based approaches, which focus on replaying episodic memory \cite{robins1995catastrophic}, and the core of which is to select representative samples or features from historical data \cite{rebuffi2017icarl,aljundi2019gradient,jihwan2021rainbow}.
\textit{(ii)} Distributed memory representation approaches \cite{fernando2017pathnet,mallya2018packnet}, which allocate individual networks for specific knowledge to avoid interference, represented by Progressive Neural Networks (PNN) \cite{rusu2016progressive}.
While such methods can well-address the catastrophic forgetting issue, they face the parameter explosion problem. \textit{(iii)} Synaptic memory approaches, which optimize the gradient toward the defined direction by regularization. \cite{Kirkpatrick2016Overcoming,aljundi2018memory,zenke2017continual,pan2020continual} selectively retained weights that are highly relevant to the previously learned knowledge. \cite{zeng2019continual,tang2021layerwise} kept the parameter space of the preceding and following tasks orthogonal to avoid interference. Such methods do not rely on storing additional data or parameters but face the gradient error accumulation effect.
Unlike the above mechanisms, we introduce a new memory consolidation mechanism, i.e., learning knowledge through a short-term memory network and then transferring the short-term memory to long-term memory.

\textbf{Memory transfer in lifelong learning.}
Unlike transfer learning, memory transfer is concerned with preserving previously learned knowledge while facilitating the learning of new knowledge. This makes traditional transfer learning methods inapplicable in the lifelong learning setting \cite{ke2020continual}. Current work on memory transfer focus on distributed memory representations for overcoming catastrophic forgetting, e.g., \cite{rusu2016progressive} based on the transferability of features in neural networks, \cite{yosinski_2014_NIPS} transferring features from previous models. \cite{rajasegaran2019random} reused specific modules of the historical model and extends new modules to learn new knowledge. \cite{abraham2019plasticity,cong2020gan} adapted old parameters to the new task by extra adaption networks. Nevertheless, to the extent of our knowledge, no current approaches consider adverse effects of memory transfer with a conflict between old and new knowledge. In particular, these approaches ignore the hindrances of old knowledge to the learning of new knowledge. In contrast, this work aims to handle the anterograde forgetting issue through individual networks equipped with transferring cells, and overcomes the catastrophic forgetting issue through a hybrid strategy of synaptic memory and long- and short-term memory transformation.

\textbf{Distributed frameworks for lifelong learning.} Several works start to focus on distributed frameworks for lifelong learning \cite{delange2021continual}, typically represented by \cite{rusu2016progressive}, which assign a separate network to each task. \cite{ostapenko2019learning,rajasegaran2019random,yoon2018lifelong,li2019learn} proposed task-adaptive dynamic networks, and \cite{schwarz2018progress} built memory-learning coupled frameworks to separately learn and store knowledge. Unlike these approaches, our CMN does not allocates distinct sub-networks for each task but encodes long- and short-term networks based on the cycle of memory. Although \cite{kemker2018fearnet,delange2021continual} also introduced the concept of long- and short-term memory, we use a memory consolidation process which distills and transfers condensed knowledge from the short-term memory network to the long-term memory network.

Similar to our work, P$\&$C \cite{schwarz2018progress} distills new knowledge into the knowledge base via two separate networks, but there are some differences to be clarified, \textit{(i)} CMN designs a novel module called transfer cell to transfer memory precisely; in contrast, P$\&$C transfer memory by lateral connection as the same as PNN, without considering the negative transfer; \textit{(ii)} CMN utilizes the function-based regularization to preserve memory rather than the parameter-based one in P$\&$C.

\begin{figure*}[t]
\centering
\includegraphics[width=.8\linewidth]{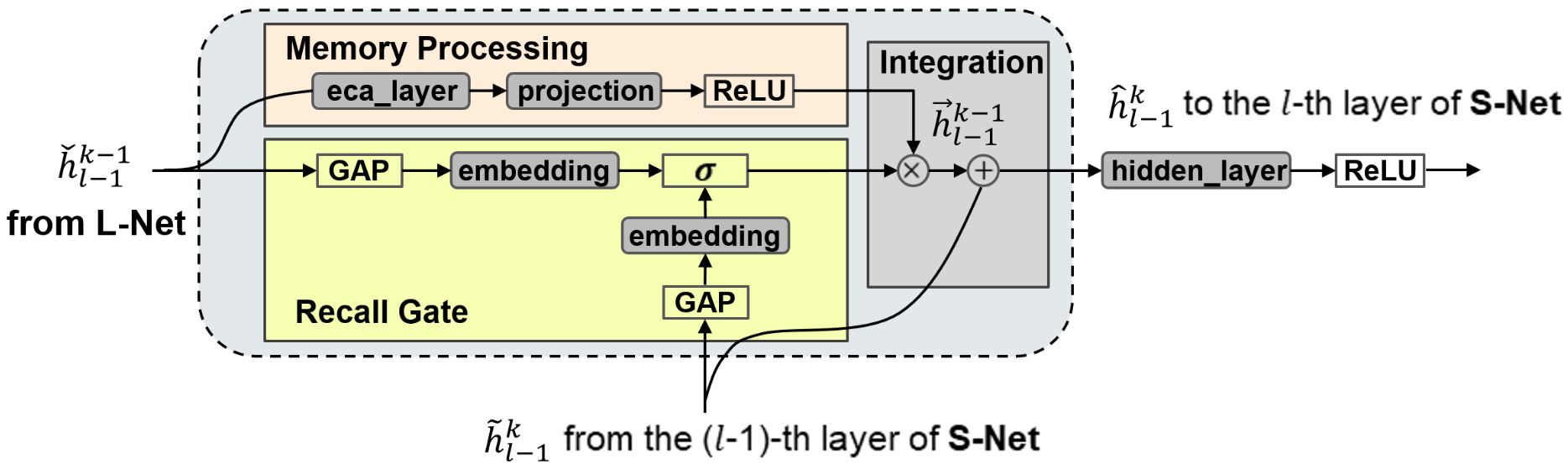}
\caption{Structure of the transfer cell.It consist of four branches:memory processing, recall gate, and memory integration.}
\label{fig3}
\end{figure*}

\section{Method}

\subsection{Overview of Cycled Memory Architecture}
Since the anterograde forgetting issue comes from two parts: \textit{(i)} capacity shrinkage due to memory retention and \textit{(ii)} conceptual confusion due to irrelevant memory migration, and motivated by the functional organization \cite{alberini2009transcription,izquierdo1999separate} in the mammalian brain(stated in principle \textit{(i)} and \textit{(ii)}, Introduction), we proposed the separate architecture called CMN, which contains two sub-networks: a Long-Term Memory Network (L-Net) and a Short-Term Memory network (S-Net), as shown in Fig. \ref{fig2}, to store old knowledge and learn new knowledge, respectively. The S-Net is independent of the L-Net to avoid the network's capacity for the new task being occupied by the previous task; the transfer cell learns to recall and extract relevant features from the long-term memory to supplement the short-term memory(L-S transfer half-cycle). Besides, a memory consolidation strategy(corresponding to principle \textit{(iii)}) is developed to allow the integration of new knowledge learned in the short-term memory network into the long-term memory network (S-L transform half-cycle). Combined with memory consolidation based on knowledge distillation, CMN achieves the cycle of long-term memory - short term memory - long term memory and constructs a sound framework for lifelong learning (corresponding to lifelong learning requirements).


\textbf{Denotation preview.} For simplicity, we consider a sequence of $T$ classification tasks, in which the $k$-th task contains ${C^k}$ classes. We aim to train a lifelong learning model which can learn the $k$-th task without forgetting and being disturbed by the old knowledge learned in previous ${k-1}$ tasks. We set the two networks the same architecture except for the output layer, which is not necessary. Specifically, when it learns the $k$-th task, the output layer of L-Net contains $\sum_{i=1}^{k-1} C^{i}$ neurons for predicting all previous $k-1$ tasks; accordingly, S-Net is only responsible for learning the knowledge of the current task, so its output layer contains only ${C^k}$ neurons. Given a training data set for the $k$-th task $\mathcal{D}^{k} = \left \{ \textbf{x}_i,\textbf{y}_i \right \}$, $i=1,2,\cdots,N_k$. We denote the weights of the L-Net for the previous $k-1$ task as $\theta_L^{k-1}$ and the features of layer $l-1$ of the L-Net as  $\check{h}_{l-1}^{k-1}$. The weights of the S-Net are $\varphi_S^{k}$, and the weights of S-Net's layer $l$ are $W_{l}^{k}$. The features of S-Net's layer $l-1$ are $\tilde{h}_{l-1}^{k}$. The feature of L-Net operated by the Recall Gate is $\vec{h}_{l-1}^{k-1}$. The confluence of the transferred $\vec{h}_{l-1}^{k-1}$ and $\tilde{h}_{l-1}^{k}$ after the Integration module is $\hat{h}_{l-1}^{k}$.

\textbf{L-S transfer half-cycle.} Fig. \ref{fig2}(a) shows the process of long-term memory facilitates learning new tasks via transfer cells. Before training the S-Net, CMN freezes L-Net's weights $\theta_L^{k-1}$ and initializes the S-Net's weights $\varphi_S^{k}$. The features $\check{h}_{l-1}^{k-1}$ of the L-Net flow into the S-Net through the transfer cell and merge with features $\tilde{h}_{l-1}^{k}$ of S-Net. The S-Net is trained by adjusting $\varphi_S^{k-1}$ only to minimize the following cross-entropy loss:
\begin{equation}\label{eq:1}
{\varphi_S^{k}}^{*} = \argmax \limits_{\varphi_S^{k}}\left \{ -\frac{1}{N_k}\sum_{i=1}^{N_{k}} \mathbf{y}_{i} \cdot \log  \left(f(\mathbf{x}_{i};\varphi_S^{k}, \theta_L^{k-1}) \right) \right \}
\end{equation}
where $\cdot$ denotes the inner product and $f(\mathbf{x}_{i};\varphi_S^{k}, \theta_L^{k-1})$ denotes the output of S-Net parameterized by $\varphi_S^{k}$ for a given input $\mathbf{x}_{i}$ with transferred features from L-Net parameterized by $\theta_L^{k-1}$. The feature transfer from L-Net to S-Net is enabled by a transfer cell, which is detailed in Section 3.2.

\textbf{S-L transform half-cycle.} Next, in the memory consolidation phase, shown in Fig. \ref{fig2}(b), CMN freezes the weights $\varphi_S^{k}$ of S-Net and the L-Net's weights $\theta_L^{k-1}$ for previous  $k-1$ tasks before training the L-Net. To integrate the knowledge of historical knowledge and new knowledge, two distillation losses are utilized. One is to preserve the memory of L-Net, and the other is to transform the short memory to the L-Net. They adjusts $\theta_L^{k-1}$ to ensure the distribution of the new L-Net consistent with the one combining both old L-Net for previous $k-1$ tasks and new S-Net for the current $k$-th task.

\begin{table*}[htbp]
  \small
  \caption{Comparison of $AF$ on class-incremental and cross-domain learning}
  \label{tabel1}
  \centering
  \setlength{\tabcolsep}{0.85pt}{
  \begin{tabular}{@{}ccccccccccc@{}}
  \toprule
  $AF(\%)$          & tasks                 & One & Joint                             & fine & PNN      & LwF      & EWC      & DGR       & Hnet     & CMN      \\ \midrule
  \multirow{2}{*}{CIFAR10}  & 2   & \multirow{2}{*}{\textbackslash{}}   & \multirow{2}{*}{\textbackslash{}} &  $ -23.51\pm2.33 $ & $\textbf{20.71}^{2}\pm1.68 $    & $-23.22\pm1.76$   & $7.44\pm2.54$    &  $-15.86\pm1.94$  & $\textbf{8.50}^{3}\pm 2.12$  & $\textbf{26.33}^{1}\pm2.05 $   \\
                          & 5   &    &     & $-7.25\pm1.10$  & $\textbf{18.14}^{2}\pm 3.13$    & $-64.59\pm 11.34$ & $-5.09\pm 2.22$ & $-55.53\pm6.24$ & $\textbf{7.36}^{3}\pm3.03$ &$\textbf{20.24}^{1}\pm 2.90$ \\ \cmidrule(r){1-11}
  \multirow{4}{*}{CIFAR100} & 2   & \multirow{4}{*}{\textbackslash{}}  & \multirow{4}{*}{\textbackslash{}} &  $-21.02\pm 0.75$  & $\textbf{17.38}^{1}\pm2.80$ & $\textbf{-9.31}^{3}\pm 0.87$  & $-31.71\pm1.58$  & $-39.01\pm13.18$   & $-73.66\pm5.12$ & $\textbf{11.64}^{2}\pm9.48$    \\
                          & 5   &   &   & $-35.99\pm1.10$   & $\textbf{22.70}^{1}\pm2.26$ & $-46.87\pm 0.38$ & $\textbf{-9.88}^{3}\pm 1.35$ & $-40.45\pm 5.79$ & $-21.52\pm 2.75$ & $\textbf{10.45}^{2}\pm 0.69$    \\
                          & 10  &    &   &$-32.95\pm 1.67$& $\textbf{24.60}^{2}\pm0.93$ &$-54.57\pm 11.32$& $\textbf{-15.21}^{3}\pm 4.75$ & $-72.95\pm7.52$  & $-37.88\pm0.32$ & $\textbf{26.40}^{1}\pm3.80$ \\
                          & 20  &    &      & $-25.58\pm 0.63$ & $\textbf{29.31}^{2}\pm1.21$ & $-75.29\pm0.36$ & $\textbf{-18.99}^{3}\pm 2.77$ & $-94.87\pm13.50$ & $-21.05\pm 0.30$ & $\textbf{30.39}^{1}\pm 0.43$ \\ \cmidrule(r){1-11}
  \multirow{2}{*}{Cal/VOC}   & C-V &  \multirow{2}{*}{\textbackslash{}}   & \multirow{2}{*}{\textbackslash{}} &$\textbf{-12.85}^{3}\pm 3.45$ & $\textbf{13.61}^{2}\pm 1.48$ & $-45.06\pm 2.71$ & $-25.91\pm 2.02$ & $-49.91\pm 11.68$   &  $-23.42\pm 5.81$   & $\textbf{18.70}^{1}\pm 4.57$ \\
                          & V-C &    &    & $\textbf{-9.70}^{3}\pm 0.54$ & $\textbf{15.66}^{2}\pm 3.96$ & $-55.98\pm6.51$& $-18.76\pm 1.51$ &     $-24.14\pm 9.23$     & $-46.69\pm 0.91$ & $\textbf{17.96}^{1}\pm 2.47$ \\ \bottomrule
  \end{tabular}}
\end{table*}

\subsection{Transfer Cell}
The distinct transfer cell is deployed between the corresponding layers of the L-Net and the S-Net (Fig. \ref{fig3}). It is consist of three primary modules: \textit{(i)} memory processing which maps the L-Net memory into the same space as S-Net, \textit{(ii)} retrieval gate, which produces a value ranging from 0 to 1 to control the amount of knowledge to be transferred, and \textit{(iii)} knowledge integration which combines the S/L-Nets’ knowledge using multiplications and additions.
In summary, features of L-Net flow to the S-Net through the transfer cell, in the order of Memory processing--Recall gate--Memory integration. For illustration, assume that the S-Net learns the $k$-th task without transfer cell, the activation of the $l$-th layer is formulated as:
\begin{equation}\label{eq:2}
\tilde{h}_{l}^{k}= \textbf{ReLU}(W_{l}^{k}\tilde{h}_{l-1}^{k})
\end{equation}
where $\textbf{ReLU}$ is the activation function.

\textbf{Memory processing.} It processes the feature $\check{h}_{l-1}^{k-1}$ of L-Net. It functions in two ways: (1) the channel attention module \cite{wang2020eca}, denoted as $eca\_layer_{l-1}^{k-1}$, is used to learn high-relevant features; (2) the projection layer $P_{l-1}^{k-1}$, is used to map the output of $eca\_layer_{l-1}^{k-1}$ to the same space as the features of S-Net to achieve alignment. The output $\vec{h}_{l-1}^{k-1}$ is calculated by
\begin{equation}\label{eq:6}
\vec{h}_{l-1}^{k-1} = \textbf{ReLU}(P_{l-1}^{k-1}eca\_layer_{l-1}^{k-1}(\bar{h}_{l-1}^{k-1}))
\end{equation}

\textbf{Recall Gate.} It calculates the recall gate $g_{l-1}^{k-1}$, controlling the knowledge from the long-term memory network to be recalled for short-term memory formation. The recall gate embeds the features extracted from L-Net and S-Net into a common low-dimensional space, and learns the feature transferability between long- and short-term memories. $g_{l-1}^{k-1}$ can be calculated by,
\begin{equation}\label{eq:5}
g_{l-1}^{k-1} = \sigma(\bar{\textbf{E}}_{l-1}^{k-1}{GAP}(\check{h}_{l-1}^{k-1})+ \tilde{\textbf{E}}_{l-1}^{k}{GAP}(\tilde{h}_{l-1}^{k})+ \textbf{b}_{l-1}^{k-1:k} )
\end{equation}
where $\sigma(\cdot )$ is the sigmoid function used to normalize the values to $\left ( 0,1 \right ) $ . $\bar{\textbf{E}}_{l-1}^{k-1}$ and $\tilde{\textbf{E}}_{l-1}^{k}$ are linear projection layers to transform features to a common subspace, corresponding to L-Net and S-Net. $\textbf{b}_{l-1}^{k-1:k}$ is bias item. In addition, to reduce the computational cost of the linear layer, the Global Average Pooling (GAP) is utilized to reduce the dimension of the features $\check{h}_{l-1}^{k-1}$ and $\tilde{h}_{l-1}^{k}$.

\textbf{Memory integration.} It recalls the processing feature $\vec{h}_{l-1}^{k-1}$ from long-term memory combining with the recall gate, and integrates it with feature $\tilde{h}_{l-1}^{k}$ of short-term memory. It calculates $\hat{h} _{l-1}^{k}$ by
\begin{equation}\label{eq:4}
\hat{h} _{l-1}^{k} = \tilde{h}_{l-1}^{k}+  g_{l-1}^{k-1}\otimes \vec{h}_{l-1}^{k-1}
\end{equation}
where $g_{l-1}^{k-1}$ is the recall gate calculated by the memory recall branch, $\vec{h}_{l-1}^{k-1}$ is the post-processing on the original feature $\check{h}_{l-1}^{k-1}$ via the memory processing branch, and $\otimes$ is the element-wise multiplication operator.

Correspondingly, the activation in the $l$-th layer of S-Net with transfer cell is processed by
\begin{equation}\label{eq:3}
\tilde{h}_{l}^{k}= \textbf{ReLU}(W_{l}^{k}\hat{h} _{l-1}^{k})
\end{equation}
where $\hat{h} _{l-1}^{k}$ is the output of memory integration branch.

In particular, we found the initialization of the linear layer is significant and best initialization is constant 1 (details in Fig. \ref{fig10}, Appendix). Moreover, the results in Fig. \ref{fig11}(Appendix) demonstrate that using Batch Normalization \cite{ioffe2015batch} does not improve the speed and accuracy of training, and even has a negative impact. We also discuss the transfer strategy in Fig. \ref{fig9}, Appendix, and exhibit that the transfer cell based transfer strategy is the most effective method on the case of negative transfer and positive transfer.

\subsection{Memory Consolidation}
The memory consolidation uses two knowledge distillation losses to preserve and integrate long- and short-term memories. To preserve the long-term memory, we first want to keep the output distribution of new L-Net $\theta_L ^{k}$ and the output distribution of previous L-Net $\theta_L ^{k-1}$ consistent in the same dimension on $\mathcal{D}^{k}$. The objective function is defined as
\begin{equation}\label{eq:7}
L_{dis\_long}^{k}=cross\_entropy(\sigma (\frac{f(\textbf{X}^k;\theta_L^{k}) }{T} ),\sigma (\frac{f(\textbf{X}^k;\theta_L^{k-1}) }{T} ))
\end{equation}
where $T$ is a hyper-parameter to control the concentration of the output distribution; the larger $T$ is, the more divergent the distribution is, and vice versa.

For integrating the short-term memory into the L-Net, similar to \cite{hinton2015distilling}, we define a hybrid objective function as
\begin{equation}\label{eq:8}
\begin{aligned}
L_{dis\_short}^{k}=\left ( 1-\beta \right )cross\_entropy\left (  \sigma(f(\textbf{X}^k;\theta_L^{k}) ),\textbf{Y}^k\right ) \\ +  \beta T^{2}cross\_entropy\left ( \sigma (\frac{f(\textbf{X}^k;\theta_L^{k}) }{T} ),\sigma (\frac{f(\textbf{X}^k;\varphi_L^{k}) }{T} ) \right )
\end{aligned}
\end{equation}
where $\beta$ is a hyper-parameter to control the degree of knowledge integration from S-Net. The larger value means more knowledge of S-Net is integrated in the new L-Net.

In order to integrate S-Net and L-Net, the following total objective function is minimized to update L-Net:
\begin{equation}\label{eq:9}
L_{distill}^{k} = L_{dis\_long}^{k}+  L_{dis\_short}^{k}
\end{equation}

\section{Experiments and Results}

\subsection{Experimental Settings}

\textbf{Overview of experiments, dataset, models, and training details.} We implemented two experimental scenarios to investigate (1) the case of task relations known, including two situations. \textbf{a.} Task conflict, i.e., sequentially learn noise data and CIFAR-10 \cite{Krizhevsky2009Learning}, to evaluate the ability to resist irrelevant knowledge, and \textbf{b.} excluding the above case, i.e., sequentially learn CIFAR-10 and STL-10 \cite{coates2011analysis}, to evaluate the ability to resist capacity shrinkage. (2) The case of task relations unknown, including class-incremental learning on CIFAR-100 \cite{Krizhevsky2009Learning} and cross-domain sequential learning on PASCAL VOC2007 \cite{everingham2010pascal} and Caltech-101 \cite{Fei-Fei2006One-shot}, to evaluate the overall performance of continual learning. In addition, the experiments in Fig. \ref{fig1} were re-conducted to verify CMN's ability of overcoming anterograde forgetting.
Furthermore, in Appendix, we describe \textit{(i)} the data and pre-processing in Section Details of dataset and pre-processing; \textit{(ii)} Basic classification networks and \textit{(iii)} Training details.

%

\textbf{Metrics.} For a comprehensive assessment of the algorithm's performance to overcome anterograde forgetting, we designed a new metric called $AF$(details in appendix), which evaluates the cumulative average of the performance decline of a single task in the task sequence during continual learning.
It summarizes the cumulative accuracy gap of the continual learning model on the current task relative to a single-task model learning the same task and the cumulative accuracy gap of the continual learning model across all seen tasks to a multi-task model learning the same tasks jointly.

Furthermore,
we utilized the evaluation criteria of \cite{lopez2017gradient}. Specifically, the average accuracy ($ACC$) was used to evaluate the overall performance of the approach on a sequence of tasks. The forward transfer ($FWT$) was used to evaluate the ability of the approach to transfer memorized knowledge to facilitate the learning of new tasks.
Moreover, the backward transfer ($BWT$) was utilized to evaluate the ability to overcome forgetting previous tasks' knowledge. In addition, we utilized the $Params$ to evaluate the size of the model parameter and designed the $\textit { Iteration time }$ to evaluate the data storage (details in Table. \ref{tabel5}, Appendix).

\begin{figure}[t]
\centering
\includegraphics[width=\linewidth]{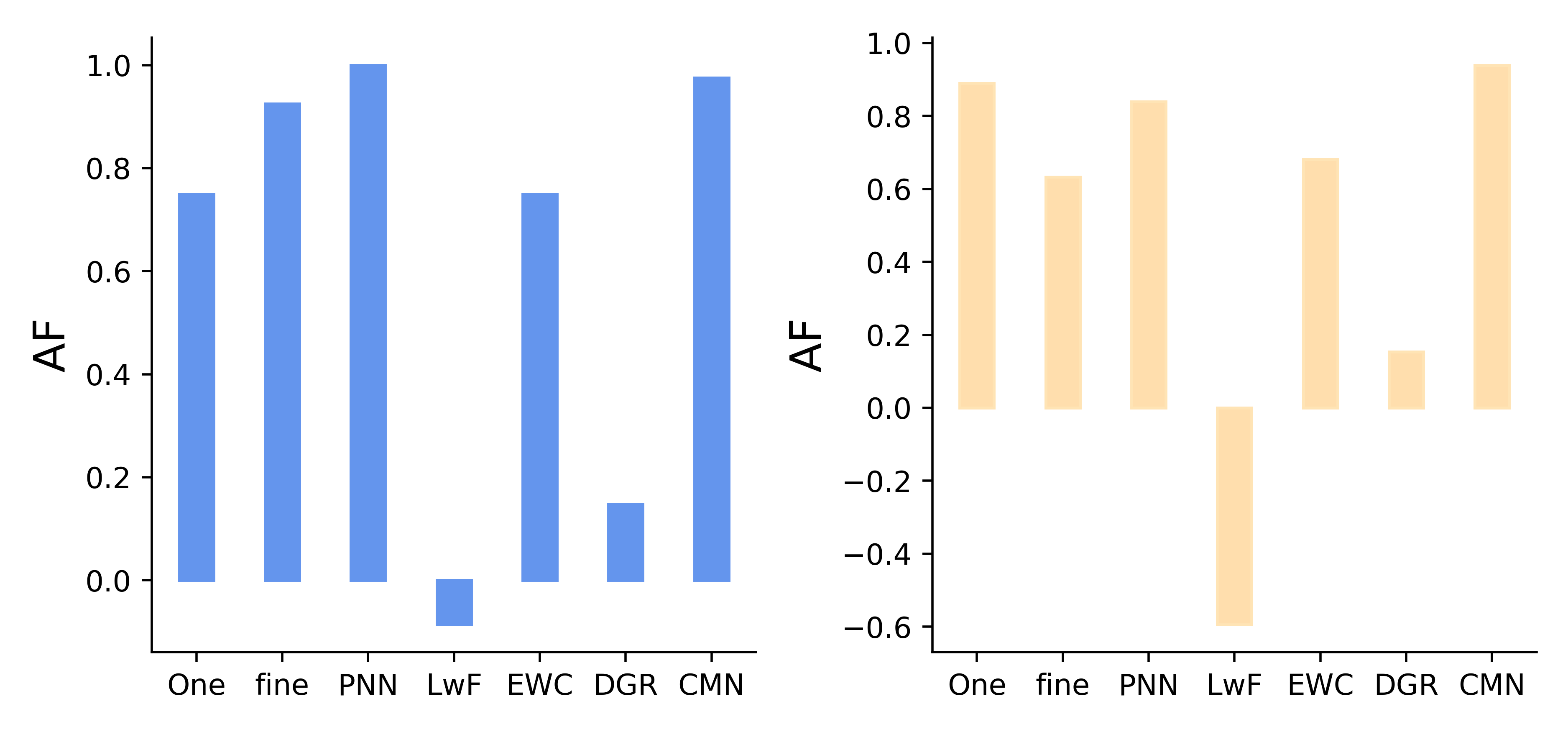}
\caption{Performance on task-related\textit{(left)} and task-conflict\textit{(right)}. \textit{left}: Results of transferring knowledge to STL-10, and \textit{right}: results of transferring noise to CIFAR-10.}
\label{fig4}
\end{figure}

\begin{figure}[t]
\centering
\includegraphics[width=\linewidth]{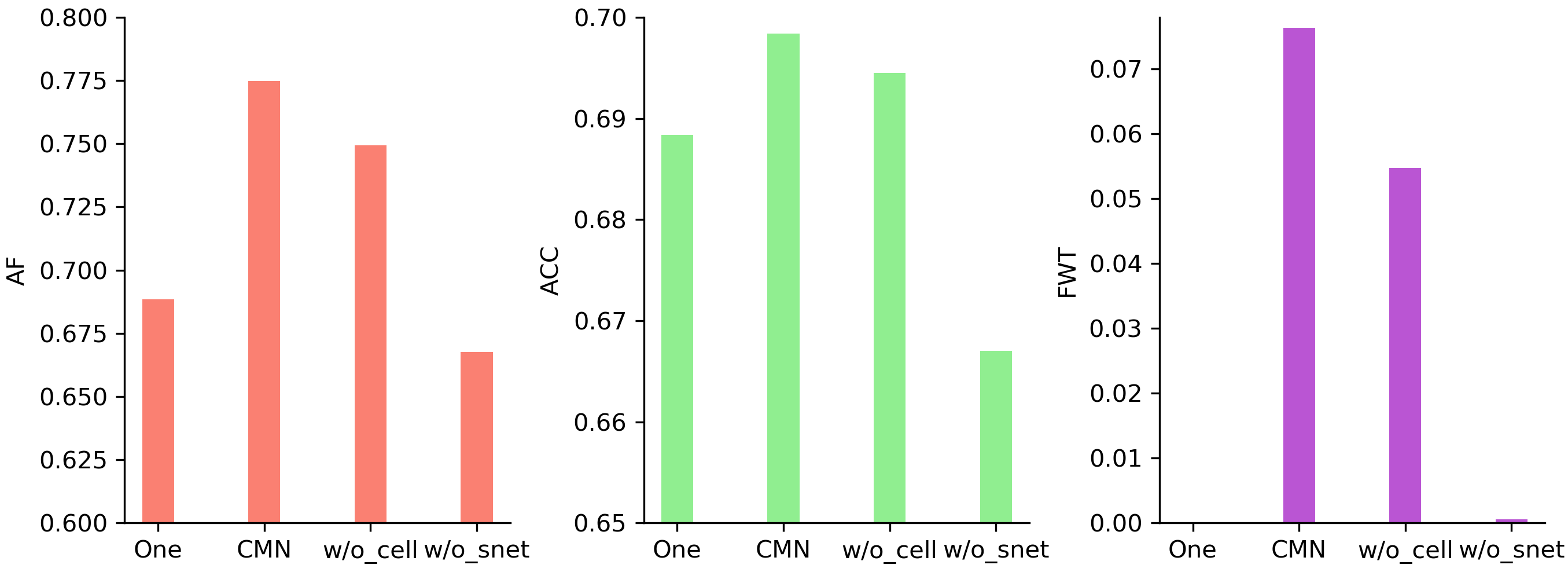}
\caption{Ablation analysis for S-Net and transfer cells on $AF$, $ACC$, and $FWT$.}
\label{fig5}
\end{figure}

\begin{figure}[t]
\centering
\includegraphics[width=0.95\linewidth]{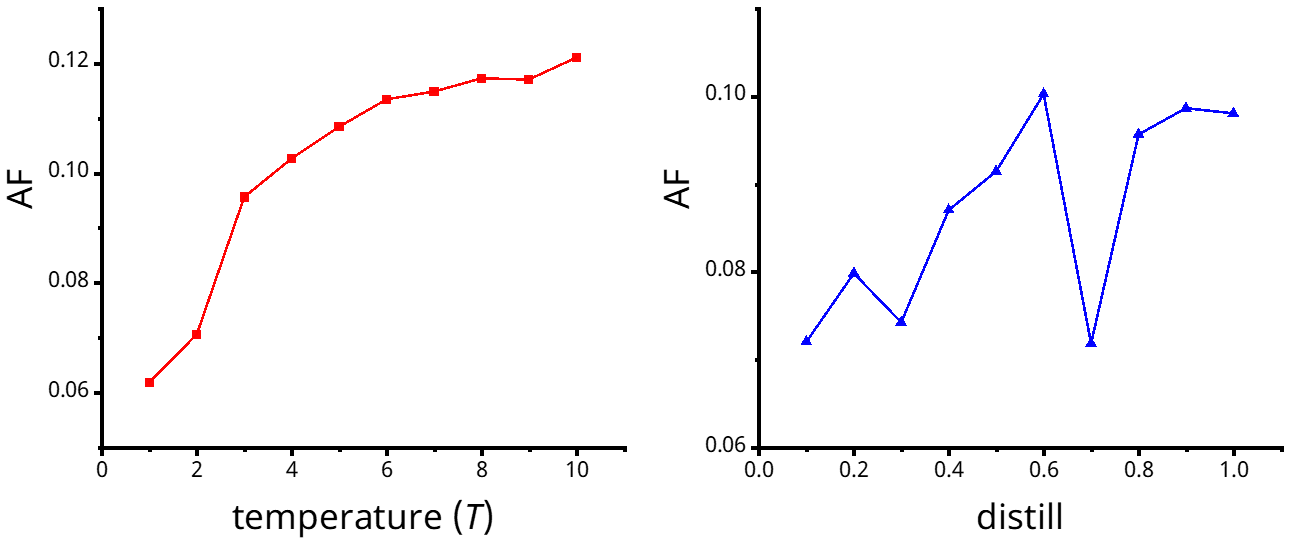}
\caption{Robust analysis of hyper-parameters $T$ and $\beta$.}
\label{fig6}
\end{figure}

\begin{figure*}[!htbp]
\centering
\includegraphics[width=0.9\linewidth]{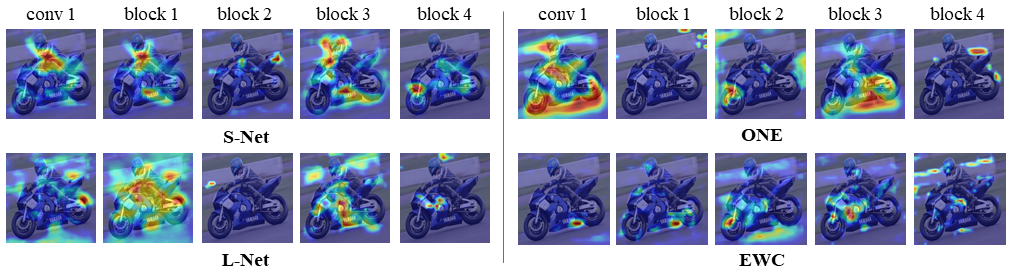}
\caption{Visualization of features on conv1 and four blocks for CMN, One, and EWC.}
\label{fig7}
\end{figure*}

\textbf{Baseline.} We compare our proposed CMN with state of the art approaches, including experience rehearsal-based method: DGR \cite{shin2017continual}; distributed memory-based: PNN \cite{rusu2016progressive}, hnet \cite{oshg2019hypercl}; and synaptic memory-based method: EWC \cite{Kirkpatrick2016Overcoming}, LwF \cite{li2017learning}. We also compared with some traditional methods, including One (individual training model for each task), Joint (jointly training all tasks at once), fine-tuning (fine-tuning on the previous model without constraints).

\subsection{Results}


\textbf{Class-incremental and cross-domain.} We first repeated the experiments in Fig. \ref{fig1} with CMN. In Fig. \ref{fig12}(Appendix), results show that CMN can significantly reduce anterograde forgetting. In memory retention (left), CMN's S-Net undergoes almost no anterograde forgetting because its network capacity for the new task is constant; although CMN's L-Net cannot wholly overcome anterograde forgetting, it still dramatically slows down anterograde forgetting compared to the EWC algorithm. The results on memory transfer (right) are also consistent with our findings.

We then compared our CMN with the above methods on standard tasks, including class-incremental and cross-domain, in Table \ref{tabel1}. The average results running five times of operation demonstrate that CMN achieves the highest score on $AF$ in most cases, significantly better than other algorithms. It achieves the second score in some cases. In Table \ref{tabel2} and Table \ref{tabel3}, Appendix, we further compared the performance on $ACC$ and $FWT$, which partly evaluate algorithms' ability to address the two cases of anterograde forgetting.  The results demonstrate that CMN performs closely to PNN in several cases and even outperforms PNN and other approaches on both $ACC$ and $FWT$ when learning long sequence of tasks. Although PNN performs close to our algorithm and even outperforms CMN in a few cases, it faces the parameter explosion problem. As shown in Table \ref{tabel5} (Appendix), PNN has a significantly higher number of parameters than the other algorithms.
Moreover, its parameter size increases rapidly with the number of tasks (Fig. \ref{fig13} in Appendix). In Fig. \ref{fig14} (Appendix), we exhibit PNN's performance declines rapidly with a constant number of parameters on $AF$, $ACC$, $BWT$. In contrast, CMN uses only 1/5 of the testing parameters and 1/2 training parameters of PNN, but achieves or even outperforms PNN in the case of a fixed number of parameters. Note that CMN's parameters are constant instead of dramatically increasing with tasks' length like PNN. In addition, compared the \textit{Iteration time} with the baseline, CMN costs less time compared with DGR but achieves better performance.

We future found that CMN may benefit from the separate network architecture. The confusion matrix in Fig. \ref{fig8} (Appendix) exhibits that CMN preserves a higher accuracy on the earlier learned categories than EWC when learning categories increase. The quantitative result in Table \ref{tabel4} (Appendix) supports this conclusion. CMN achieves a dramatic score on $BWT$, closing to the usual upper limit PNN.


\textbf{Task-related and task-conflict.} Fig. \ref{fig4}\textit{(left)} compares these methods in task-related cases, where a learning model is sequentially trained on CIFAR-10 and STL-10.
On $AF$, PNN and CMN obtain the best results compared to fine-tuning, a typical transfer learning approach. The fine-tuning, PNN, and CMN exceed One by around 15\%. 
Fig. \ref{fig4}\textit{(right)} compares the performance of these methods in resisting irrelevant knowledge transfer. We assumed the extreme case of transferring knowledge from noisy data to the learning of CIFAR-10. The results show that CMN is the only method among these methods to address the negative migration effect of irrelevant memory. CMN ranks first on $AF$ and is the only method with a positive value on $FWT$. Compared to CMN, all other baselines on the new task are disturbed by the introduction of noise.


\subsection{Analysis}
\textbf{Component ablation analysis.} To investigate the contribution of each component of the CMN, we performed ablation experiments on CIFAR-100 with the first two tasks for two crucial components: the use of short-term memory network and the transfer cell. Fig. \ref{fig5} presents the results of algorithms on $AF$, $ACC$, and $FWT$ after removing different components. It indicates that both short-time memory network and transfer cells enhance the model's performance, but the former contributes more. The model's performance sharply drops when removing the short-term memory network, even below One, i.e.,11.0\% for $AF$, 3.2\% for $ACC$, and 7.8\% for $FWT$. The model's performance decreases when removing the transfer cell but is still higher than One, i.e., $AF$ decreases by 2.5\%, $ACC$ decreases by 0.5\%, and $FWT$ decreases by 2.2\%.

\textbf{Stable to hyper-parameters.} Fig.\ref{fig6} investigates the proposed approach's stability for two significant hyper-parameters, beta distill $\beta$ and temperature $T$, in various ranges. The $\beta$ controls the amount of knowledge transferred from S-Let to L-Net, and the larger its value, the more knowledge is distilled from S-Net into L-Net. The $T$ controls the temperature of knowledge distillation, and the larger its value, the more divergent the distribution of distilled soft label is.
The two parameters collaborate to reach a balance between the old and new memories. The results indicate that the model is tolerant to $T$. The model's performance increases with the increase of $T$. In contrast, the model is more sensitive to the value of $\beta$, which reaches the best performance at a value of 0.6.  Overall, it is acceptable, the $AF$ floats within 3\% when $\beta$ varies in the range of $\left [ 0.1, 1 \right ] $.

\textbf{Feature learning visualization.} We analyzed the features of various layers of networks on Caltech-101 and compared them with One and EWC by Grad-CAM \cite{selvaraju2017grad}. In Fig. \ref{fig7}, compared with One and EWC, CMN grasps more representative characteristics, especially on the higher level. For example, on block 4, S-Net and L-Net capture features in a few critical local regions to represent the concept of a motorcycle. In contrast, One learns fewer local vital features, and EWC learns features of the most discrete areas. S-Net learns richer features than L-Net, mainly because L-Net selectively integrates features as it needs to balance old and new knowledge when integrating short-term memory. Furthermore, this phenomenon is not coincidental, and we obtained consistent findings on other samples, as shown in Fig. \ref{fig15}(Appendix).



\section{Conclusion}
Learning and memory are the core of lifelong learning, one essential step towards an open world intelligence. However, in the past years, we ignored the interference effect of memory on learning, i.e., anterograde forgetting issue. Inspired by the human brain's memory mechanisms, this paper suggests the designing principle for the functional architecture of lifelong learning. In particular, we propose a cycled memory network to overcome the anterograde forgetting. It transforms short-term memory to long-term memory through a memory consolidation mechanism and transfers long-term memory to learning through transfer cells. Comprehensive experiments have been conducted to verify the effectiveness of CMN.
While our results show that using two individual short-term and long-term memory networks can effectively address the anterograde forgetting, this architecture could be further explored. Besides, the proposed framework focuses on the classification task and does not consider the network's evolution strategy when long-time memory is saturated. It is of great significance to explore ways to address these issues by drawing on the brain's memory mechanisms.
In future work, we will explore the properties and corresponding structures of different memories and the engram of memory storage, such as meta-knowledge. In addition, we will consider integrating the mechanisms about synaptic memory and sleep consolidation memory in neuroscience into the proposed memory-learning framework.

\bibliography{aaai22.bib}

\clearpage

\appendix

\section{Basic classification networks}
Considering the complexity of tasks, we utilized the resnet-34 \cite{he2016deep} in the cross-domain experiments considering, and resnet-18 \cite{he2016deep} in other experiments. Besides, we utilized similar architectures for L-Net and S-Net, except for the different units in the output layer.

\section{Training details}
All experiments used the optimization strategy of stochastic gradient descent with an initialized learning rate $\left \{ 1,0.1, 0.001 \right \} $. The momentum was 0.9, and weight decay was $1\times 10^{-5}$. The grid search was utilized to get the optimal result. The training epoch was 100. Besides, all models' weights were initialized with $kaiming\_uniform$. The mini-batch size was set to 54 in the experiment of cross-domain and set to 1024 in other experiments. In Appendix, we details the hyper-parameters setting and computational resources in Table. \ref{tabel6} and Computational resources.

\begin{figure}[t]
\centering
\includegraphics[width=0.95\linewidth]{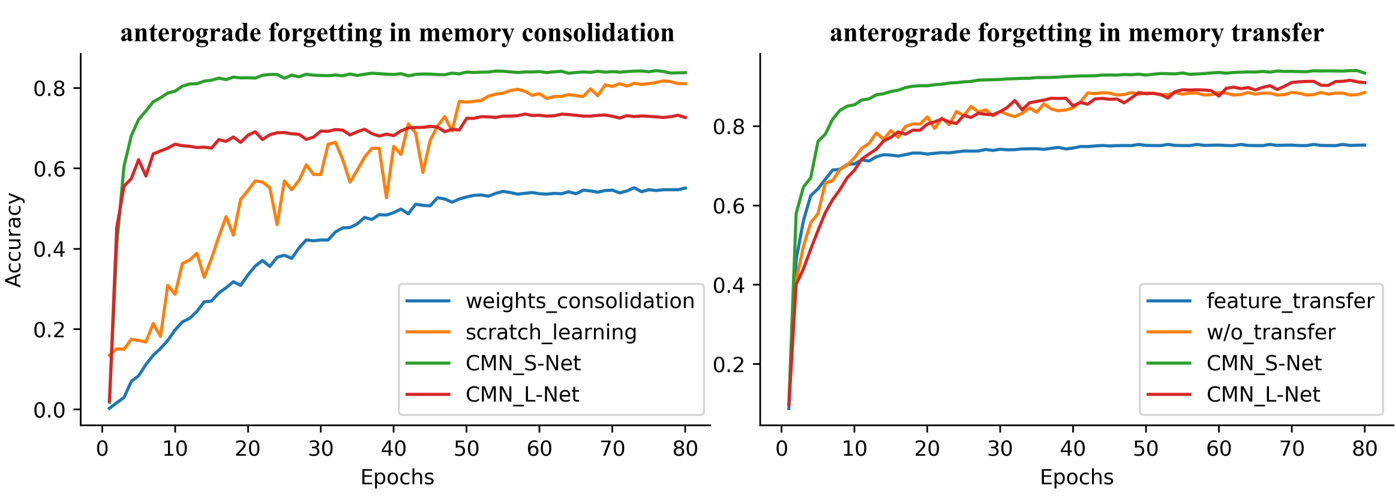}
\caption{Overcome anterograde forgetting with CMN corresponding to the experiments in Fig. 1.}
\label{fig12}
\end{figure}

\begin{figure}[t]
\centering
\includegraphics[width=0.95\linewidth]{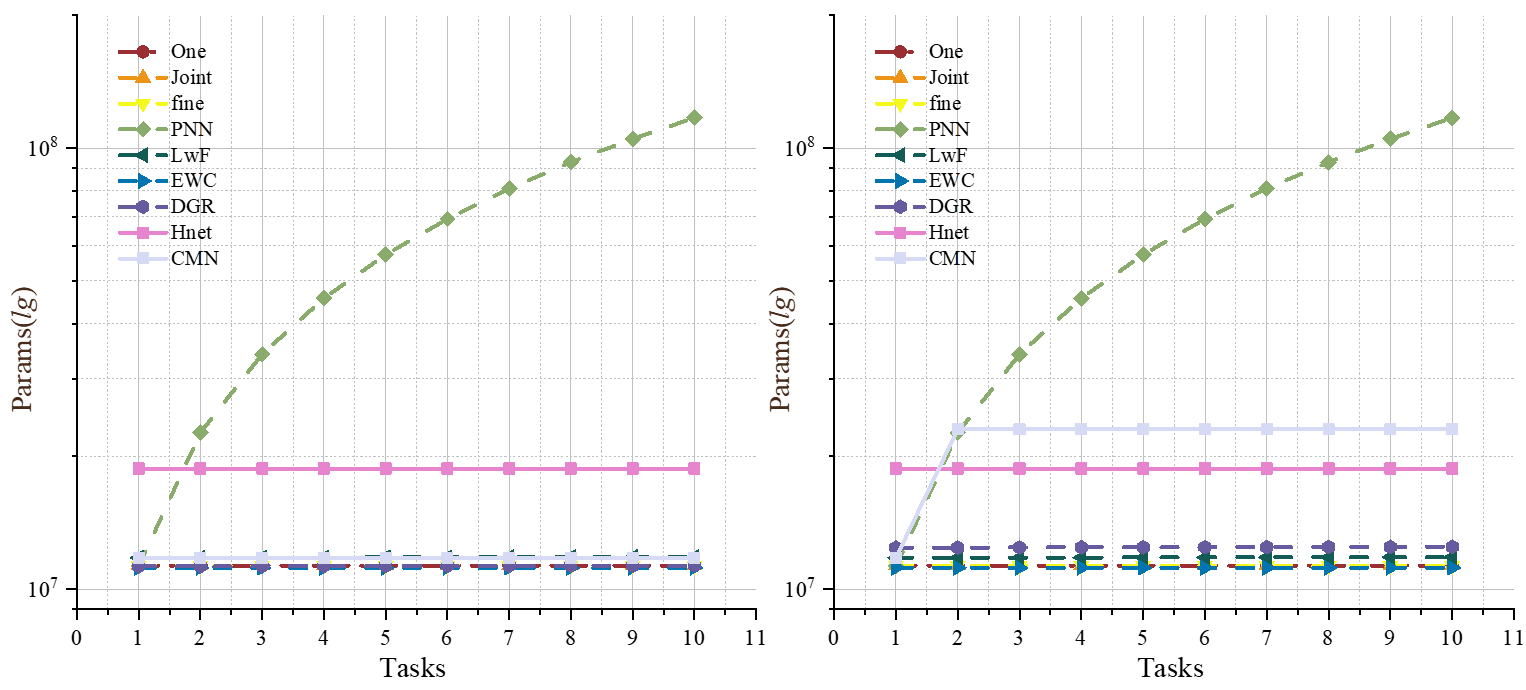}
\caption{Comparison of testing and training parameter size with tasks increase. \textit{left}, testing Params comparison and \textit{right}, training Params comparison}
\label{fig13}
\end{figure}

\begin{figure}[h]
\centering
\includegraphics[width=0.95\linewidth]{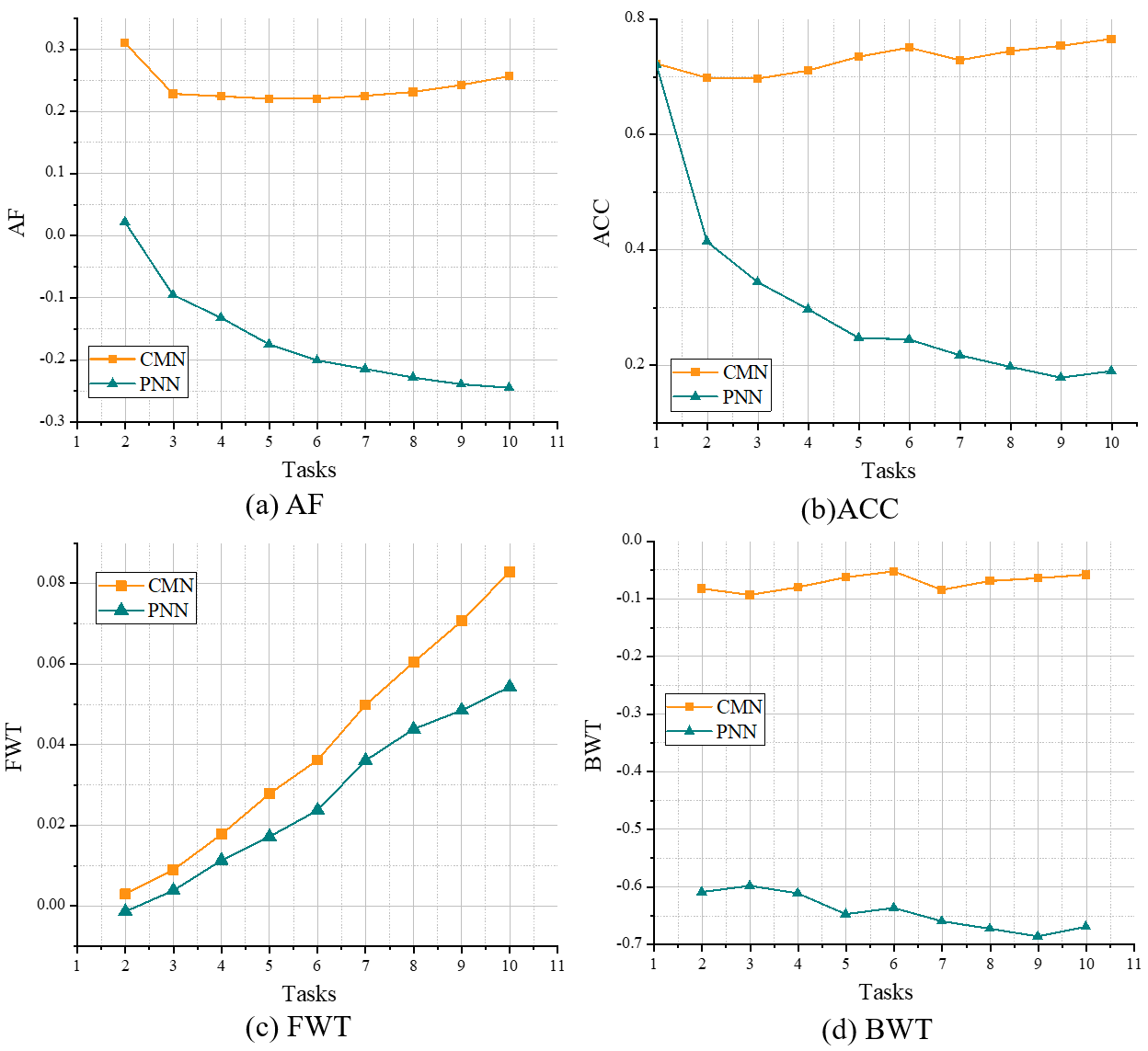}
\caption{Comparison between CMN and PNN under identical parameter size setting}
\label{fig14}
\end{figure}

\begin{table*}[t]
  \small
  \caption{Comparison of $ACC$ on class-incremental and cross-domain learning}
  \label{tabel2}
  \centering
  \setlength{\tabcolsep}{0.8pt}{
  \begin{tabular}{@{}ccccccccccc@{}}
  \toprule
  $ACC(\%)$          & tasks                 & One & Joint                             & fine & PNN      & LwF      & EWC      & DGR       & Hnet     & CMN      \\ \midrule
  \multirow{2}{*}{CIFAR10}  & 2   & $\textbf{88.6}^{3}\pm 2.31$  & \multirow{2}{*}{$68.9\pm 0.7$} &  $ 47.91 \pm 0.22 $ & $\textbf{89.19}^{2} \pm 0.32 $    & $61.68 \pm 4.30$   & $75.75 \pm 0.62$    &  $64.60 \pm 4.24$ & $87.58 \pm 0.60$  & $\textbf{91.20 }^{1}\pm 0.14 $   \\
                          & 5   &  $\textbf{92.8}^{3} \pm 4.60$  &     & $67.23 \pm 0.41$  & $\textbf{93.30}^{2} \pm 0.19$    & $ 34.38 \pm 3.14 $ & $64.78 \pm 6.00 $ & $25.06 \pm 2.56$& $88.56 \pm 3.17$ &$\textbf{94.43}^{1} \pm 0.30 $ \\ \cmidrule(r){1-11}
  \multirow{4}{*}{CIFAR100} & 2   & $\textbf{68.9}^{3}\pm 0.54$  & \multirow{2}{*}{$57.0\pm 0.7$} &  $ 35.45 \pm 0.24 $ & $\textbf{71.53}^{1} \pm 1.45 $    & $55.63 \pm 0.35$   & $37.60 \pm 1.30$    &  $47.64 \pm 3.58$  & $28.61 \pm 1.02$  & $\textbf{69.30}^{2} \pm 0.33 $   \\
  & 5   &  $\textbf{62.5}^{3} \pm 3.58$  &     & $15.63 \pm 0.53$  & $\textbf{70.51}^{1} \pm 0.48$    & $ 27.18 \pm 0.52$ & $38.02 \pm 0.48$ & $16.1 \pm 2.28$& $49.55\pm 1.04$ &$\textbf{63.87}^{2} \pm 0.24 $   \\
                          & 10  &  $\textbf{75.2}^{3} \pm 1.50$  &   & $16.82 \pm 0.36$ & $\textbf{78.00}^{1} \pm 2.02$ &$25.01\pm 5.02$ & $37.32 \pm 4.78$ & $7.37 \pm 0.67$  & $43.77 \pm 0.34$ & $\textbf{77.86}^{2}\pm 3.46$ \\
                          & 20  &  $\textbf{84.3}^{1} \pm 0.21$  &      & $21.62\pm 0.26$ & $\textbf{82.33}^{2} \pm 0.38$ & $16.46\pm 0.28$ & $37.05\pm 6.38$ & $2.90 \pm 0.52$ & $53.56\pm 0.35$ & $\textbf{82.29}^{3} \pm 0.37$  \\ \cmidrule(r){1-11}
  \multirow{2}{*}{Cal/VOC}   & C-V &  \multirow{2}{*}{$\textbf{45.0}^{3}\pm 4.34$}   & \multirow{2}{*}{$43.0\pm 0.3$} &$22.26\pm 0.56$ & $\textbf{49.75}^{1} \pm 0.90$ & $30.64 \pm 0.18$ & $17.98\pm 0.32$ & $11.25 \pm 3.79$   &  $25.59\pm 6.21$   & $\textbf{45.90 }^{2}\pm 4.34$ \\
                          & V-C &    &    & $32.61\pm 0.53$ & $\textbf{55.58}^{1}\pm 2.86$ & $29.53\pm 3.61$& $28.86\pm 0.32$ &     $27.24\pm 6.61$    & $20.77\pm 0.24$ & $\textbf{53.98}^{2}\pm 2.58$ \\ \bottomrule
  \end{tabular}}
\end{table*}

\begin{table*}[!htbp]
  \small
  \caption{Comparison of $FWT$ on class-incremental and cross-domain learning}
  \label{tabel3}
  \centering
  \setlength{\tabcolsep}{1.2pt}{
  \begin{tabular}{@{}ccccccccccc@{}}
  \toprule
  $FWT(\%)$          & tasks                 & One & Joint                             & fine & PNN      & LwF      & EWC      & DGR       & Hnet     & CMN      \\ \midrule
  \multirow{2}{*}{CIFAR10}  & 2   & 0   & \multirow{2}{*}{\textbackslash{}} &  $ -2.55\pm 1.78 $ & $\textbf{0.40}^{3}\pm1.46 $    & $-11.40\pm 4.71 $   & $\textbf{0.56 }^{2}\pm 1.67$    &  $-11.59 \pm 3.09$  & $-10.21\pm 1.73 $  & $\textbf{4.0}^{1} \pm 1.77 $   \\
                          & 5   &   0 &     & $-2.18 \pm 1.08$  & $\textbf{2.35}^{2} \pm 5.19$    & $-27.84 \pm 10.94$ & $\textbf{0.75}^{3}\pm 3.40$ & $-17.93 \pm 6.89$ & $-4.41 \pm 2.77$ &$\textbf{3.77}^{1}\pm 5.15$ \\ \cmidrule(r){1-11}
  \multirow{4}{*}{CIFAR100} & 2   & 0  & \multirow{4}{*}{\textbackslash{}} &  $\textbf{0.53}^{3} \pm 0.78$  & $\textbf{2.85}^{2} \pm 1.73$     & $-7.94 \pm 0.36$  & $-12.3 \pm 0.90$  & $-29.64 \pm 9.95$   & $-41.83\pm 2.05$ & $\textbf{4.50}^{1} \pm 1.91$    \\
                          & 5   & 0  &   & $-0.41 \pm 0.32$   & $\textbf{12.51}^{1} \pm 4.32$ & $-23.03 \pm 0.41$ & $\textbf{3.69}^{3}\pm 1.16$ & $-14.50 \pm 4.88$ & $-11.28 \pm 0.97$ & $\textbf{7.93}^{2} \pm 0.18$    \\
                          & 10  &  0  &   &$-0.038\pm 0.48$& $\textbf{4.62}^{2} \pm 0.34$ & $-34.41\pm 7.73$ & $\textbf{0.32}^{3}\pm3.40$ & $-36.53 \pm 4.59$   & $-26.16\pm 0.16$ & $\textbf{8.44}^{1}\pm 0.94$ \\
                          & 20  &   0 &      & $\textbf{0.006}^{3}\pm 0.34$ & $\textbf{2.18}^{2} \pm 0.27$ & $-49.65 \pm 0.36$ & $-3.13\pm 0.97$ & $-51.31 \pm 15.85$ & $-22.00 \pm 0.17$ & $\textbf{4.68}^{1} \pm 0.30$ \\ \cmidrule(r){1-11}
  \multirow{2}{*}{Cal/VOC}   & C-V &  \multirow{2}{*}{0}   & \multirow{2}{*}{\textbackslash{}} &$\textbf{7.92}^{2}\pm 2.94$ & $\textbf{6.88}^{3}\pm 1.72$ & $-32.67\pm 2.56$ & $-0.87\pm 2.08$ & $-18.14\pm 8.33$   &  $-5.99\pm 7.08$   & $\textbf{16.22}^{1}\pm3.63$ \\
                          & V-C &    &    & $\textbf{0.61}^{3}\pm 0.38$ & $\textbf{3.00}^{2}\pm 1.02$ & $-42.59\pm 2.75$ & $-4.69\pm 0.70$ &   $-8.35\pm 5.90$  & $-24.55\pm 0.79$ & $\textbf{6.90}^{1}\pm 0.54$ \\ \bottomrule
  \end{tabular}}
\end{table*}

\begin{table*}[!htbp]
  \small
  \caption{Comparison of $BWT$ on class-incremental and cross-domain learning}
  \label{tabel4}
  \centering
  \setlength{\tabcolsep}{1.2pt}{
  \begin{tabular}{@{}ccccccccccc@{}}
  \toprule
  $AF(\%)$          & tasks                 & One & Joint                             & fine & PNN      & LwF      & EWC      & DGR       & Hnet     & CMN      \\ \midrule
  \multirow{2}{*}{CIFAR10}  & 2   & \multirow{2}{*}{\textbackslash{}}   & \multirow{2}{*}{\textbackslash{}} &  $ -23.51 \pm 2.33 $ & $\textbf{20.71}^{2} \pm 1.68 $    & $-23.22 \pm 1.76$   & $7.44 \pm 2.54$    &  $-15.86 \pm 1.94$  & $\textbf{8.50}^{3}\pm 2.12$  & $\textbf{26.33}^{1} \pm 2.05 $   \\
                          & 5   &    &     & $-7.25 \pm 1.10$  & $\textbf{18.14}^{2}\pm 3.13$    & $-64.59\pm 11.34$ & $-5.09\pm 2.22$ & $-55.53 \pm 6.24$ & $\textbf{7.36}^{3} \pm 3.03$ &$\textbf{20.24}^{1}\pm 2.90$ \\ \cmidrule(r){1-11}
  \multirow{4}{*}{CIFAR100} & 2   & \multirow{4}{*}{\textbackslash{}}  & \multirow{4}{*}{\textbackslash{}} &  $-21.02\pm 0.75$  & $\textbf{17.38}^{1} \pm 2.80$ & $\textbf{-9.31}^{3}\pm 0.87$  & $-31.71 \pm 1.58$  & $-39.01 \pm 13.18$   & $-73.66 \pm 5.12$ & $\textbf{11.64}^{2} \pm 9.48$    \\
                          & 5   &   &   & $-35.99 \pm 1.10$   & $\textbf{22.70}^{1} \pm 2.26$ & $-46.87\pm 0.38$ & $\textbf{-9.88}^{3}\pm 1.35$ & $-40.45\pm 5.79$ & $-21.52\pm 2.75$ & $\textbf{10.45}^{2}\pm 0.69$    \\
                          & 10  &    &   &$-32.95\pm 1.67$& $\textbf{24.60}^{2} \pm 0.93$ &$-54.57\pm 11.32$& $\textbf{-15.21}^{3}\pm 4.75$ & $-72.95 \pm 7.52$  & $-37.88 \pm 0.32$ & $\textbf{26.40}^{1} \pm 3.80$ \\
                          & 20  &    &      & $-25.58\pm 0.63$ & $\textbf{29.31}^{2} \pm 1.21$ & $-75.29 \pm 0.36$ & $\textbf{-18.99}^{3}\pm 2.77$ & $-94.87 \pm 13.50$ & $-21.05\pm 0.30$ & $\textbf{30.39}^{1}\pm 0.43$ \\ \cmidrule(r){1-11}
  \multirow{2}{*}{Cal/VOC}   & C-V &  \multirow{2}{*}{\textbackslash{}}   & \multirow{2}{*}{\textbackslash{}} &$\textbf{-12.85}^{3}\pm 3.45$ & $\textbf{13.61}^{2}\pm 1.48$ & $-45.06\pm 2.71$ & $-25.91\pm 2.02$ & $-49.91\pm 11.68$   &  $-23.42\pm 5.81$   & $\textbf{18.70}^{1}\pm 4.57$ \\
                          & V-C &    &    & $\textbf{-9.70}^{3}\pm 0.54$ & $\textbf{15.66}^{2}\pm 3.96$ & $-55.98 \pm 6.51$& $-18.76\pm 1.51$ &     $-24.14\pm 9.23$     & $-46.69\pm 0.91$ & $\textbf{17.96}^{1}\pm 2.47$ \\ \bottomrule
  \end{tabular}}
\end{table*}

\begin{table*}[!htbp]
  \small
  \caption{Comparison of Parameters on each methods. }
  \label{tabel5}
  \centering
  \setlength{\tabcolsep}{2mm}{
  \begin{tabular}{@{}ccccccccccc@{}}
  \toprule
           & One    & Joint        & fine & PNN      & LwF      & EWC     & DGR    & Hnet   & CMN      \\ \midrule

 $\textit{test Params}$($M$)  &    11.326  & 11.315  & 11.326 &  $\textbf{57.366}^{1}$ & $\textbf{11.829}^{3}$ & 11.188 & 11.295 &  $\textbf{18.763}^{2}$ & 11.788 \\
 $\textit{training Params}$($M$)  &    11.326  & 11.315  & 11.326 &  $\textbf{57.366}^{1}$ &11.829 & 11.188 & 12.471 &  $\textbf{18.763}^{3}$ & $\textbf{23.575}^{2} $\\
 $\textit{Iteration time}$ &     10    &      $\textbf{50}^{1}  $      &  10     &  10 &  10 &   10      & $\textbf{20}^{2}$ &  10    &  10\\ \bottomrule
  \end{tabular}}
\end{table*}

\section{Details of dataset and pre-processing}
In task-conflict experiments, we used CIFAR-10 and Gaussian noise. In task-related experiments, we used STL-10, a benchmark of few-shot learning, and resized the sample as the same as CIFAR-10. In class-incremental, we used CIFAR-10 and CIFAR-100. CIFAR-10 was uniformly divided into 2 and 5 sub-datasets (each sub-dataset contains 2 or 5 classes). Similarly, CIFAR-100 was divided into 2,5,10, and 20 sub-datasets. In the cross-domain experiments, considering the effect of learning order, PASCAL VOC2007 and Caltech-101 were learned from two directions, and all images were first resized to 256$\times$256, and then randomly cropped  to 224$\times$224.

\section{Definition of metric $\textit{AF}$ (anterograde forgetting)}
Given $T$ task, denoting the accuracy on $i$-th task for the model trained on the $i$-th task as $P_{i,i}$, the average accuracy for the model trained on the $i$ task as $ACC_{i}$, the accuracy on the $i$-th task through the joint learning as $n_{i,i}$, and the accuracy on the $i$-th task through the one learning as $m_{i}$. We define AF metric as
\begin{equation}\label{eq:10}
AF=\frac{1}{T-1} \sum_{i=2}^{T}\left(A C C_{i}-n_{i, i}\right)+\frac{1}{T-1} \sum_{i=2}^{T}\left(P_{i, i}-m_{i}\right)
\end{equation}

\section{Definition of metric $\textit{Iteration time}$ for data storage}
We believe that low data storage is essential in lifelong learning. However, the currently commonly used training epochs do not truly reflect the model's storage cost. We believe that calculating the number of snapshots of the model observation samples is more representative of the data storage required to train the model. For instance, although all algorithms in the experiment have the same training epoch settings, the number of sample snapshots for DGR and Joint differs from the other methods because they require replaying historical data.
Denoting the number of images in a single training session as $N_{p}$, $\textit{Iteration time}$ is the number of batches needed to complete one epoch, and can be defined as
\begin{equation}\label{eq:11}
\textit { Iteration time }=N_{p} / \text { batchsize }
\end{equation}
A larger values of $\textit { Iteration time }$ indicates a higher consumption of resources in a single training epoch. For example, we have 5 tasks with 20 classes, and each class has 500 training images. And the $batchsize$ of all methods are 1024.
The $\textit { Iteration time }$ is $500 * 20 / 1024$.

We compares the model parameters size and $\textit { Iteration time }$ in 5 tasks with 20 classes per task on CIFAR100 experiments, as shown in Table. \ref{tabel5}.

\begin{figure*}[t]
\centering
\includegraphics[width=0.95\linewidth]{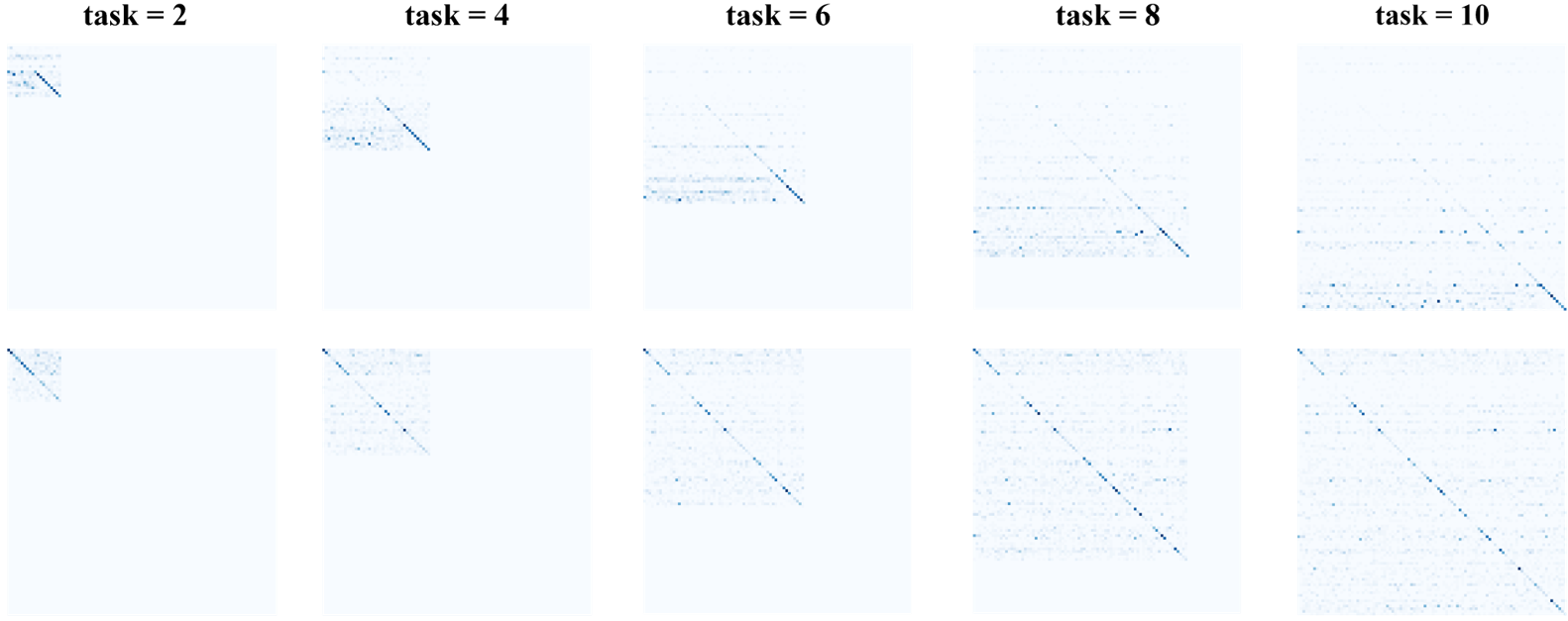}
\caption{Visualization of confusion matrix on CIFAR-100 with task num of 10. (\textit{top}): The result of EWC. (\textit{bottom}): The result of CMN.}
\label{fig8}
\end{figure*}

\section{Comparison between CMN and PNN under identical parameter size setting}
Fig. \ref{fig14} shows CMN and PNN in 10 tasks with 10 classes for CIFAR100 experiments. All methods in this part of the experiment use ResNet-18 as the backbone network.

\section{Visualization of precision confusion matrix}
Fig. \ref{fig8} investigate the ability to overcome catastrophic forgetting. We find that this advantage becomes more significant as the number of learning tasks increases. At task number 5, CMN achieves higher accuracy than EWC on the previously learned categories. Moreover, this advantage dramatically increases when the number of tasks is 10. CMN significantly performs higher accuracy on previous classes, while EWC can barely identify the previous classes.

\begin{figure}[t]
\centering
\includegraphics[width=.85\linewidth]{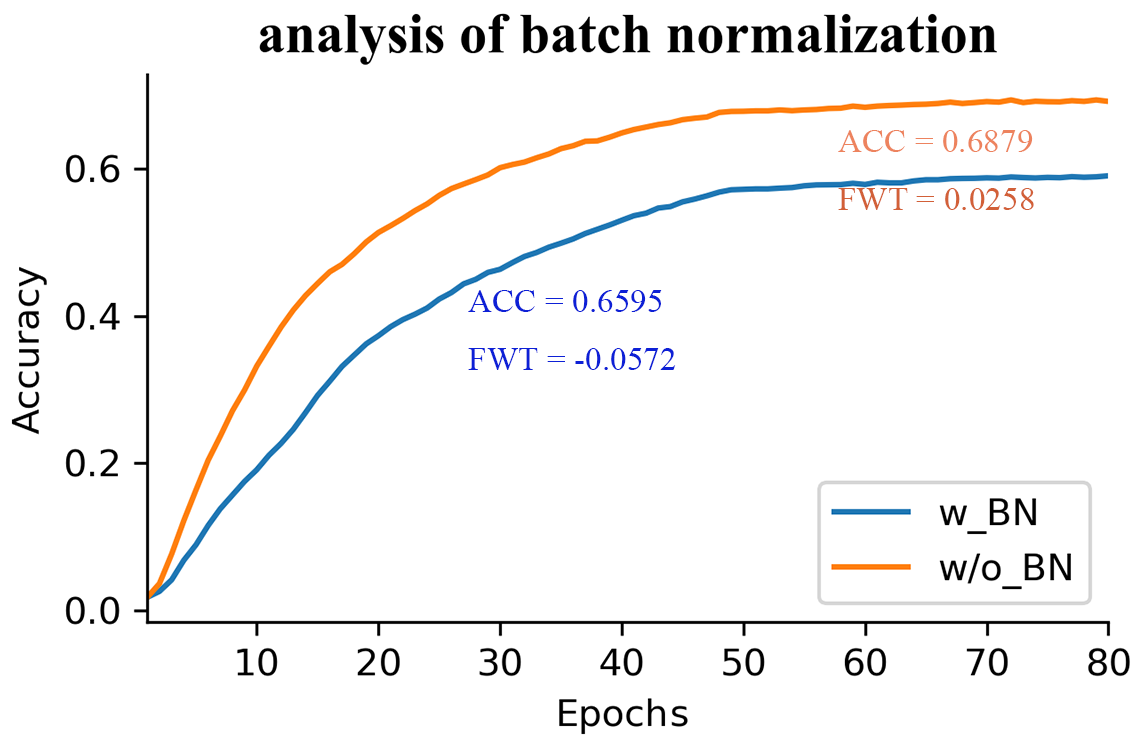}
\caption{Comparison of transfer cells with batch normalization.}
\label{fig11}
\end{figure}

\begin{figure}[t]
\centering
\includegraphics[width=.83\linewidth]{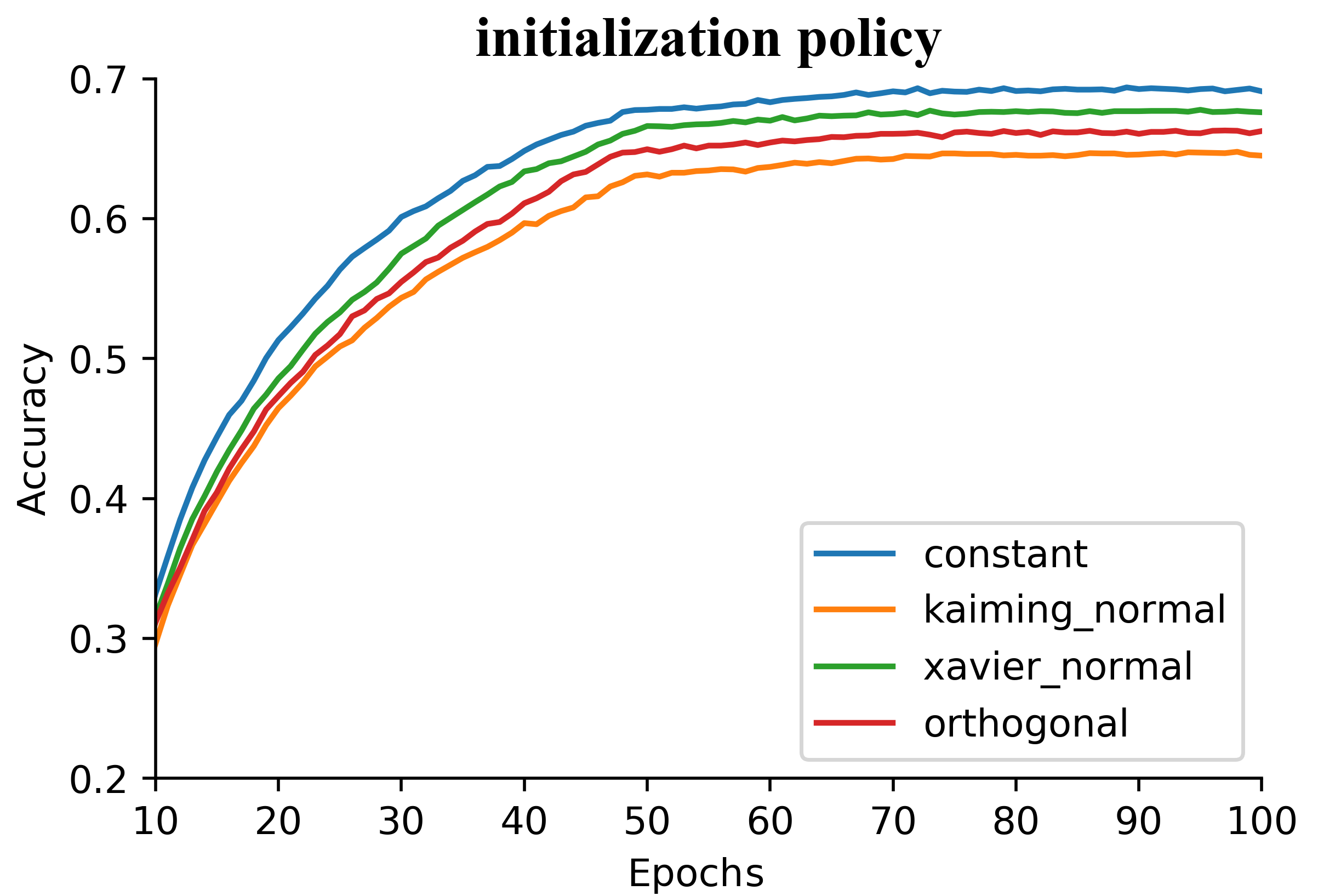}
\caption{Analysis of weights initialization for forgetting gate.}
\label{fig10}
\end{figure}

\section{Utilization of batch normalization after the transfer cell}
We compare the effect of using batch normalization in the last layer of the transfer cell. Fig. \ref{fig11} shows the experimental results on CIFAR-100 with task number 2. It indicates that the BN layer utilization in the transfer cell reduces the convergence speed and generalization accuracy on new tasks. Compared to the w/o\_BN, the performance of the model configured with w\_BN reduces by about 3\% on $ACC$ and  8\% on $FWT$.

\begin{figure*}[t]
\centering
\includegraphics[width=.8\linewidth]{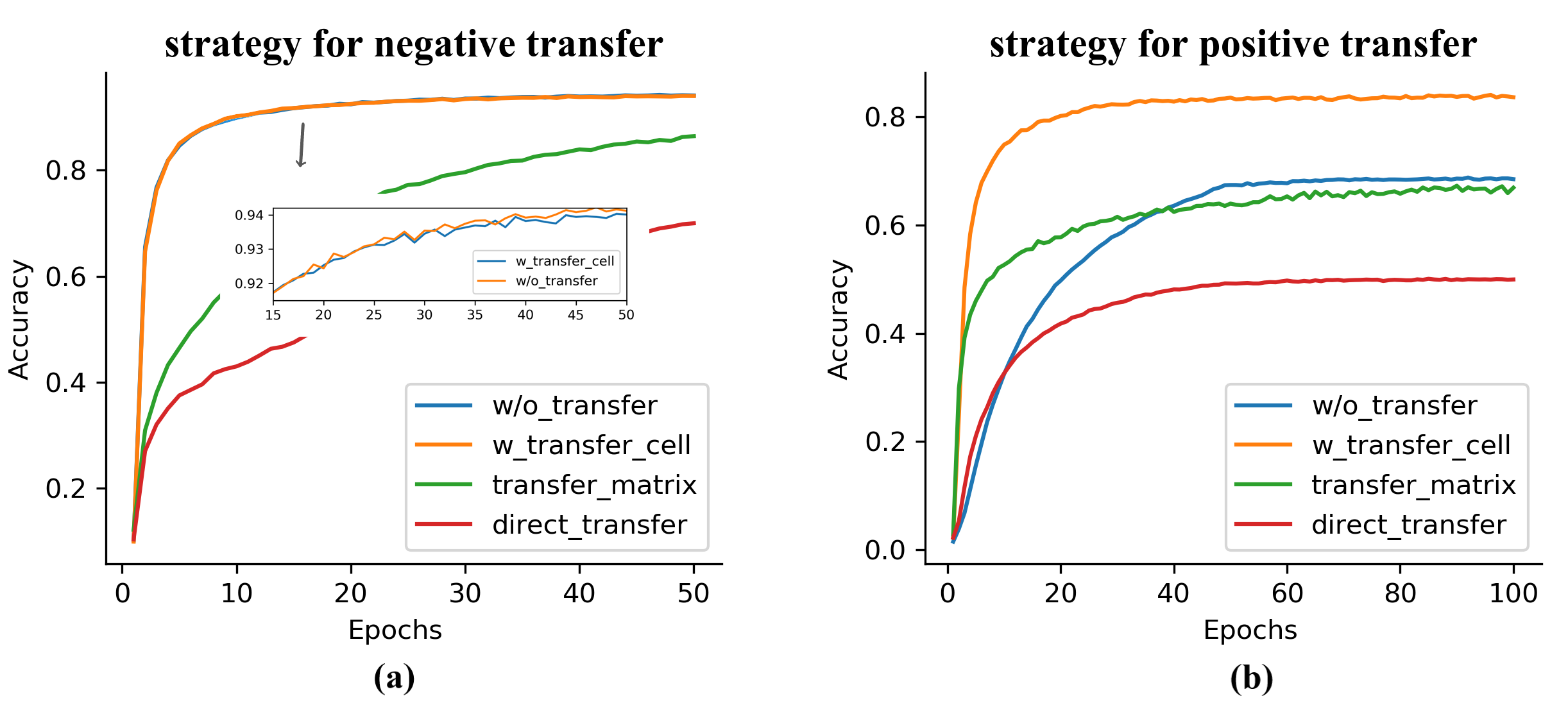}
\caption{Analysis of transfer strategy for memory transfer.}
\label{fig9}
\end{figure*}
\section{Initialization of embedding layer in the forgetting gate}
We investigate the effect of weights initialization on CMN and find that the embedding layer of the forgetting gate is susceptible to the initialization. Fig. \ref{fig10}(a) shows the model's performance on a CIFAR-100 with a task number of 2 using different initialization methods. The results indicate that it achieves the highest test accuracy and convergent fastest on the current task when the embedding layer's parameter weights are initialized to constant 1. In comparison, the accuracy is the lowest, and convergence is the slowest when using kaiming\_normal. The xavier\_normal and orthogonal are in the middle of the two. It is intuitive to leave all features to contribute equally in the initial stage during training.

\section{Transfer strategy in the transfer cell.}
We investigate the effect of transfer strategies on memory transfer. We consider two scenarios, the positive transfer case and the negative transfer case. Moreover, four transfer strategies, \textit{i} combination of transfer cell and pre-trained S-Net (w\_transfer cell); \textit{ii} using a pre-trained model on S-Net (w/o\_transfer), which does not introduce adverse transfer effects; \textit{iii} using transfer matrix (transfer\_matrix), a transfer strategy proposed by PNN; \textit{iv} directly fusing features from the same layer of L-Net (direct\_transfer), a common way of feature fusion.
Fig. \ref{fig9}(a) shows the results of negative transfer. The transfer cell avoids the negative transfer effect throughout the training phase, while neither direct\_transfer nor transfer\_matrix can handle the negative migration, especially direct\_transfer. These methods produce degradation in both accuracy and convergence speed. Fig. \ref{fig9}(b) shows the results under positive transfer. The combination of transfer cell and pre-trained model is far superior to the other strategies in terms of convergence speed and accuracy throughout the training session.  The pre-trained model approach is more effective than the feature transfer-based approach, e.g., w/o\_transfer has higher accuracy than transfer\_matrix and direct\_transfer in the late training phase. The transfer\_matrix has an 18\% higher accuracy than direct\_transfer. It indicates that the features in the two networks are not in the same space, and direct fusion has a negative impact instead.

\section{Grad CAM Visualization with CMN, EWC, and ONE}
\begin{figure*}[t]
\centering
\includegraphics[width=0.9\linewidth]{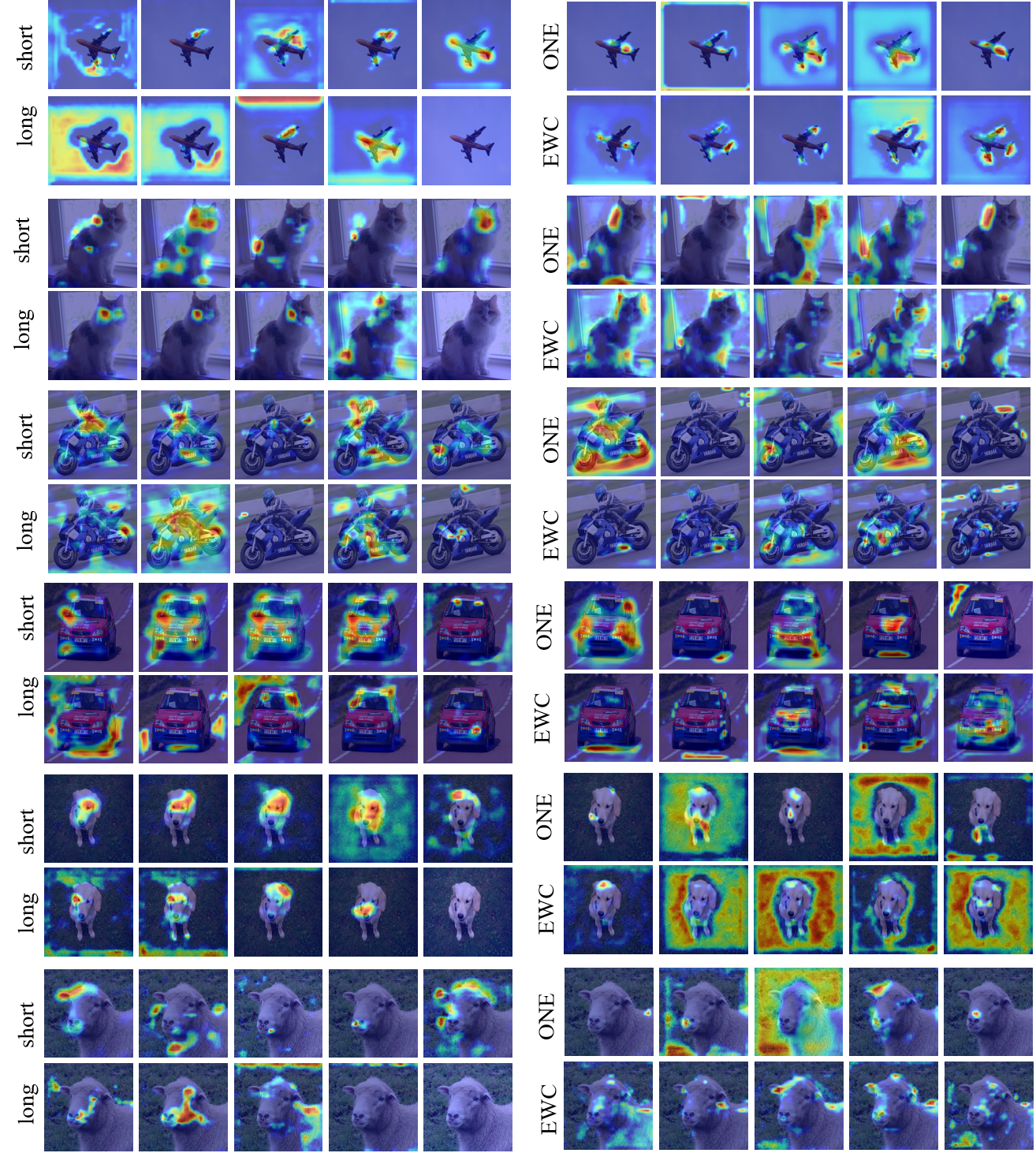}
\caption{Grad CAM Visualization with CMN, EWC, and ONE.}
\label{fig15}
\end{figure*}

We randomly selected samples from 6 classes (cat, dog, sheep, airplane, car, motorbike) from the PASCAL-VOC 2007. We chose these categories because the background of their images is simple. The results on some samples suggest that S-Net is better at learning key features than L-Net. Specifically, for block1 and conv1, the L-Net (bottom left) response is more divergent than the S-Net, which even responds to background regions. A similar phenomenon is observed at higher levels (block3), and even at block4, L-Net responds to the local region of the background. It is comprehensible because L-Net requires a trade-off between maintaining knowledge of the previous task and knowledge of the current task (although CMN largely mitigates the anterograde forgetting caused by memory retention compared to EWC). Moreover, the features learned by CMN (L-Net and S-Net) are significantly more representative than ONE and EWC. We attached the visualization of others in Fig. \ref{fig15}.

\section{Hyper-parameters of CMN on experiments.}
We list the hyper-parameters of CMN on our experiments in Table. \ref{tabel6}. Noting that, we used the uniform template $T$ value of 2, although the best $T$ is 10 in the analysis of hyper-parameters.

\begin{table}[h]
  \small
  \caption{Hyper-parameters of CMN on experiments. }
  \label{tabel6}
  \centering
  \setlength{\tabcolsep}{1.5pt}{
  \begin{tabular}{ccccccc}
\toprule
Hyper-parameters          & tasks & batch\_size         & long\_lr      & short\_lr             & beta\_distll         & temp               \\ \midrule
\multirow{2}{*}{CIFAR10}  & 2     & 1024                & 0.1                  & 0.01                  & 0.8                  & 2                  \\
                          & 5     & 1024                & 0.1                  & 0.01                  & 0.8                  & 2                  \\ \hline
\multirow{4}{*}{CIFAR100} & 2     & 1024                & 0.1                  & 0.01                  & 0.8                  & 2                  \\
                          & 5     & 1024                & 0.1                  & 0.01                  & 0.8                  & 2                  \\
                          & 10    & 1024                & 0.1                  & 0.01                  & 0.8                  & 2                  \\
                          & 20    & 1024                & 0.1                  & 0.01                  & 0.8                  & 2                  \\ \hline
\multirow{2}{*}{DOMAIN}   & C-V   & \multirow{2}{*}{54} & \multirow{2}{*}{0.1} & \multirow{2}{*}{0.01} & \multirow{2}{*}{0.8} & \multirow{2}{*}{2} \\
                          & V-C   &                     &                      &                       &                      &                    \\ \hline
few-shot                  & /       & 512                 & 0.1                  & 0.01                  & 0.8                  & 2                  \\
conflict-task             & / & 512                 & 0.1                  & 0.01                  & 0.8                  & 2                  \\ \bottomrule
\end{tabular}}
\end{table}

\section{Computational resources.}
All experiments were implemented on three deep learning stations (one is a DGX A100 Station with 4 * 80G GPU, one is with 8 * 40G A100 GPU, and one is with 2 * 48G A6000 GPU). The deep learning framework PyTorch 1.8.1 was used for the experiments.

\end{document}